\documentclass[letterpaper, 10 pt, journal, twoside]{IEEEtran}

\usepackage{graphicx}
\usepackage{color}
\usepackage{array}
\usepackage{algorithm}
\usepackage{algpseudocode}
\usepackage{amssymb}
\usepackage{soul}
\usepackage{hyperref}
\usepackage[colorinlistoftodos]{todonotes}
\usepackage{subcaption}
\usepackage{siunitx}
\usepackage{balance}

\makeatletter
\let\NAT@parse\undefined
\makeatother
\usepackage[numbers]{natbib}

\renewcommand{\baselinestretch}{1.0}

\newlength\savedwidth
\newcommand{\wcline}[1]{\noalign{\global\savedwidth\arrayrulewidth\global\arrayrulewidth 0.75pt} \cline{#1}
\noalign{\global\arrayrulewidth\savedwidth}}

\begin{document}
%
\title{VUNet: Dynamic Scene View Synthesis for Traversability Estimation using an RGB Camera}
%
%
%
\author{Noriaki Hirose, Amir Sadeghian, Fei Xia, Roberto Mart\'{i}n-Mart\'{i}n, and Silvio Savarese%
\thanks{This paper was recommended for publication by Editor Paolo Rocco upon
evaluation of the Associate Editor and Reviewers' comments. The Toyota Research Institute ("TRI")  provided funds to assist with this research, but this article solely reflects the opinions and conclusions of its authors and not TRI or any other Toyota entity. The TOYOTA Central R$\&$D Labs., INC. supported N. Hirose at Stanford University. }
\thanks{The authors are with Computer Science Department,
Stanford University, USA
{\tt\small hirose@mosk.tytlabs.co.jp}}%
}


\maketitle

\begin{abstract}
We present VUNet, a novel view(VU) synthesis method for mobile robots in dynamic environments, and its application to the estimation of future traversability. Our method predicts future images for given virtual robot velocity commands using only RGB images at previous and current time steps. The future images result from applying two types of image changes to the previous and current images: 1) changes caused by different camera pose, and 2) changes due to the motion of the dynamic obstacles. We learn to predict these two types of changes disjointly using two novel network architectures, SNet and DNet. We combine SNet and DNet to synthesize future images that we pass to our previously presented method GONet~\cite{hirose2018gonet} to estimate the traversable areas around the robot. Our quantitative and qualitative evaluation indicate that our approach for view synthesis predicts accurate future images in both static and dynamic environments. We also show that these virtual images can be used to estimate future traversability correctly. We apply our view synthesis-based traversability estimation method to two applications for assisted teleoperation.
\end{abstract}


\section{Introduction}
\IEEEPARstart{A}{utonomous} robots can benefit from the ability to predict how their actions affect their input sensor signals. The ability to predict future states provides an opportunity for taking better actions. This ability can be applied to a variety of tasks from perception and planning to safe navigation.
In robot visual navigation the actions are the velocity commands given to a robot, the input sensor signals are the images captured from the robot's RGB camera. And the predicted future images used to better understand the consequence of actions, can be predicted using scene view synthesis methods. In this context, a view synthesis model can determine which actions bring the desired sensor outcomes or can cause future hazards.  

Previous approaches have addressed the scene view synthesis problem assuming that a 3D model of the environment is available to virtually move the camera~\cite{pix2pix2017,xia2018gibson}. Recently, several approaches have relaxed this assumption and synthesized images using only a small set of previous images and a virtual action~\cite{babaeizadeh2017stochastic,flynn2016deepstereo}. However, none of these approaches can be applied to predict images for navigation in unknown environments with dynamic obstacles.
In this scenario, scene view synthesis is extremely challenging because it needs to account for both the changes in the camera pose and the motion of the dynamic obstacles.

Being able to predict the future state of both the static and the dynamic parts of an environment has multiple direct applications towards safe navigation.
One of these applications is traversability estimation: to identify traversable and non-traversable spaces in the surroundings of the robot.
Traditionally, traversability estimation methods have relied on depth sensors or on LIDARs~\cite{fox1997dynamic,flacco2012depth,pfeiffer2017perception,cinietal2002real}. However, depth sensors can fail in outdoor conditions or when the surface of the static or dynamic obstacles are reflective, and LIDARs are very expensive compared to more affordable RGB cameras, or might fail to detect glasses.
A new family of methods uses only RGB images to estimate traversability~\cite{hirose2018gonet,richter2017safe}.
However, these RGB-based methods do not have the predictive power to estimate the traversability of the locations the robot will need to navigate in the future.

\begin{figure}[t]
  \begin{center}
    \includegraphics[width=0.8\hsize]{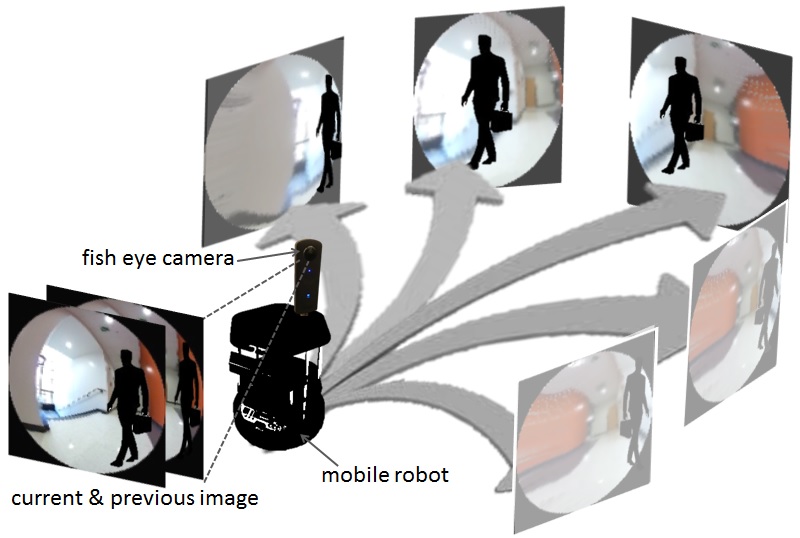}
  \end{center}
	\caption{\footnotesize{Scene view synthesis for mobile robots. Our method, VUNet predicts multiple future images assuming different navigation commands and integrating the image changes caused by changes in robot pose and by dynamic objects. The input to our method are current and previous images from an on-board fish-eye camera. The robot image is cited from \cite{turtlebot}.}}
	\label{f:intro}
	\vspace{-8mm}
\end{figure}

In this work we propose a novel deep neural network-based method for dynamic-scene view synthesis, VUNet in the context of robot navigation and its application for future traversability estimation. Our synthesis method can predict the appearance of both static (e.g., walls, windows, stairs) and dynamic (e.g., humans) elements of the environment from different camera poses in future time steps. To do that, our method requires only as input the last two acquired images and a virtual navigation command, i.e., a linear and angular velocity (Fig.~\ref{f:intro}).
We combine this method to predict future images with our previously presented RGB-based traversability estimation algorithm, GONet~\cite{hirose2018gonet}, into a system that identifies the traversable areas around the robot as well as the various velocity commands that can be executed safely by the robot.

The main contributions of this work are thus twofold: First, we propose a novel dynamic scene view synthesis method, VUNet. The technical novelty of our method is the combination of two different networks that can separately model static and dynamic transformations conditioned on robot's actions.
The proposed view synthesis method outperforms state-of-the-art methods in both static and dynamic scenes. And second, we propose a system to estimate traversability in future steps based on the synthesized images. We also propose two applications of the system in assisted teleoperation: early obstacle detection and multi-path traversability estimation.

\section{Related Work}
\label{sec:relatedwork}
We will cover in this section two main research areas: scene view synthesis, and traversability estimation.

{\bf Scene View Synthesis} is the problem of generating images of the environment from virtual camera poses. For unknown environments, a variant of the problem assumes the only input are real images taken at a certain pose. This problem has been widely studied both in computer vision~\cite{zhou2016view, ji2017deep, park2017transformation} and in computer graphics~\cite{seitz1996view, hedman2016scalable, shade1998layered} using two main types of methods. The first type of methods synthesizes pixels from an input image and a pose change with an Encoder-Decoder structure~\cite{dosovitskiy2015learning,kulkarni2015deep,tatarchenko2016multi}. The second type reuses pixels from an input image with a sampling mechanism~\cite{zhou2016view}. Instead of generating pixels, this type of method generates a flow field to morph the input image. If information from multiple views is available, a smart selection mechanism needs to be used to choose which image to sample pixels from~\cite{flynn2016deepstereo}. Previous methods focus on predicting either changes due to camera motion or due to dynamic objects~\cite{vondrick2016generating,liu2018ano_pred}, but not both.
Our method is able to deal with changes both in camera view and dynamic objects, making it suitable for dynamic scenes. 

{\bf Traversability Estimation:} Estimating which areas around the robot are safe to traverse has been traditionally done using Lidar or other depth sensors~\cite{fox1997dynamic,flacco2012depth,pfeiffer2017perception,cinietal2002real,bogoslavskyi2013efficient,cherubini2013avoiding}. 
These methods estimate the geometry of the surroundings of the robot and use it to infer the traversable areas.
However, lidar sensors are expensive and depth measurements can be affected by surface textures and materials, e.g. highly reflective surfaces and transparent objects such as mirrors and glass doors. 
These issues have motivated the use of RGB images for traversability estimation~\cite{suger2015traversability,kim2007traversability, ulrich2000appearance}. Some RGB-based methods try to first estimate depth from RGB and then apply a method based on depth images~\cite{eigen2014depth, eigen2015predicting}.
Other methods learn a generative deep neural network and formulate it as anomaly detection problems~\cite{hirose2018gonet, richter2017safe}.
For example, GONet\cite{hirose2018gonet}, which we use in our system, contains a Generative Adversarial Network (GAN) trained in a semi-supervised manner from traversable images of a fisheye camera. Since the GAN only learns to generate traversable images, GONet uses the similarity between the input image and its GAN regenerated image to estimate traversability.
GONet and other RGB-based methods estimate traversability only in the the space right next to the robot; our proposed approach predicts traversability for longer horizon trajectories.

\section{Methods}
\label{sec:model}

\subsection{Dynamic Scene View Synthesis}
\label{ssec:dsvs}

\begin{figure}[t]
\begin{center}
\hspace*{-5mm}
\begin{subfigure}[b]{0.45\textwidth}
\centering
\includegraphics[width=0.75\hsize]{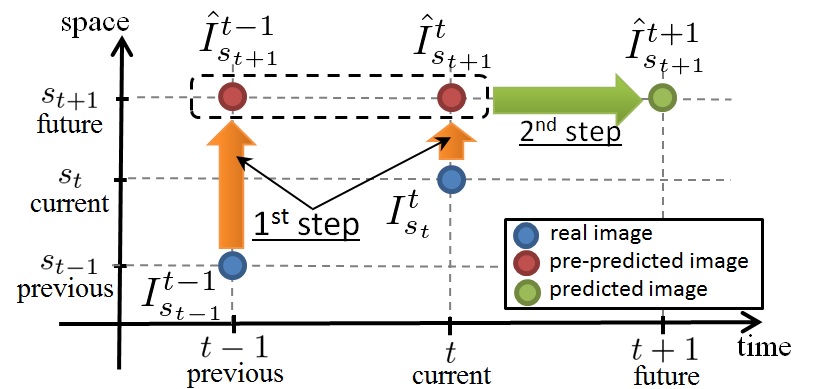}
\caption{\footnotesize{Generation of virtual images by altering both the spatial and temporal dimensions. First we generate virtual images from a different pose based on the real images (orange arrows). Then we generate a virtual image from that pose in the next time step for predicting the future position of the moving objects (green arrow). Blue circles indicate images at previous ($I_{s_{t-1}}^{t-1}$ ) and current ($I_{s_t}^{t}$) time steps.}}
\label{f:overview_net:a}
\vspace*{4mm}
\end{subfigure}
\begin{subfigure}[b]{0.45\textwidth}
\centering
\includegraphics[width=0.75\hsize]{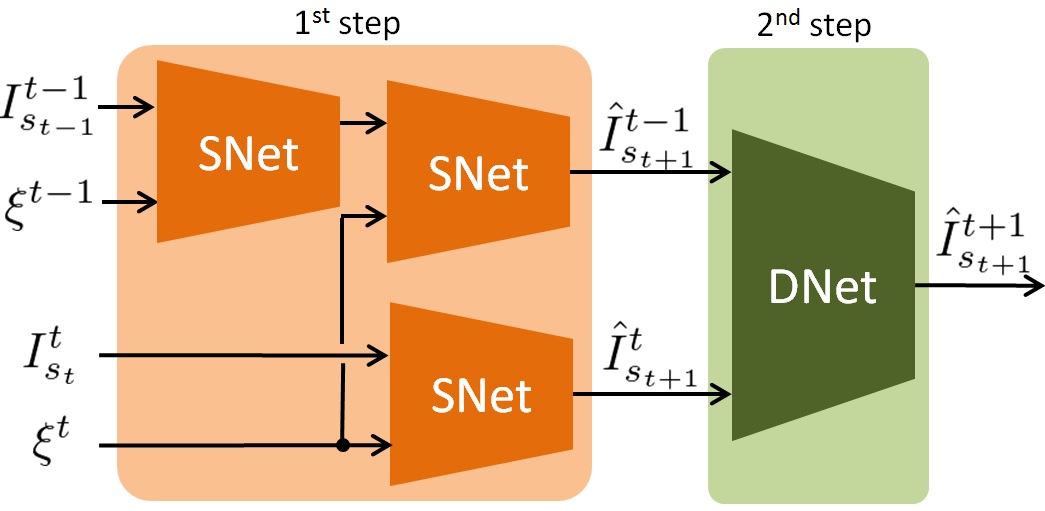}
\caption{\footnotesize{Overall network structure of VUNet for dynamic-scene view synthesis. We use SNet (twice on the previous image $I^{t-1}_{s_{t-1}}$, once on the current image $I^{t}_{s_{t}}$) to generate virtual images from the future robot pose. Then we use DNet to generate a virtual image from the future robot pose in the next time step combining changes in the static and dynamic parts of the environment}}
\label{f:overview_net:b}
\end{subfigure}
\end{center}
\caption{\footnotesize{Overview of our system, VUNet for view synthesis in dynamic environments. First step: we generate virtual images as seen from the location the robot will move to ($s_{t+1}$) at previous ($t-1$) and current ($t$) time steps using SNet. Second step: we generate a virtual image as seen from the location the robot will move to ($s_{t+1}$) at the next time step ($t+1$) using DNet over the previously generated spatially altered virtual images.}}
\label{f:overview_net}
\vspace{-4mm}
\end{figure}

In this section, we introduce VUNet for view synthesis in dynamic environments. Our goal is to generate predicted images by altering both the spatial and the temporal domains (see Fig.~\ref{f:overview_net:a}). Formally, let $\{I_{s_{t}}^{t}\}$ be a set of consecutive images, each of them captured at different time steps $t$ and (possibly) different poses ${s_{t}}$. Given this set, we aim to predict an image $I_{s_{t'}}^{t'}$ at a new robot pose $s_{t'}$ and a new time-step $t'$. Usually $t'$ will be the next time step $t+1$.
Since we are working with mobile robots the pose is parameterized by robot's position and orientation, $s_{t} = (x,y,\theta)$, $s_{t} \in \mathbb{R}^3$.
The robot command at time $t$ is a velocity in robot frame, expressed as a twist $\xi^{t} = (v^t, \omega^t)$, where $v^t$ and $\omega^t$ are the linear and angular components. We assume the mobile robot is nonholonomic and the velocity is two dimensional, $v^t, \omega^t \in \mathbb{R}$.

Our general approach is to apply changes to the last acquired real images to generate virtual images. Changes in the image are caused by two factors: changes in the viewpoint of the camera (spatial domain) and changes due to dynamically moving objects (temporal domain). 
For the static parts of the environment, the image changes are only caused by the change of viewpoint, while for dynamic objects both factors contribute to the appearance change. 
It is difficult to learn both factors simultaneously, so we propose to learn them in a disentangled manner: we completely separate the models to predict image changes in the static and moving parts of the environment, and individually train each model. 
In the following subsections we will first present our model to predict changes in the static parts of the environment due to robot motion, SNet. Then we will present the model to predict appearance changes due to motion of the dynamic parts of the environment, DNet. Finally we will explain VUNet, the combination of SNet and DNet to synthesize complete images in future time steps from different viewpoints in dynamic environments. 

\paragraph*{\textbf{Static Transformation Network (SNet)}}
\label{ssec:snet}

\begin{figure}[t]
  \begin{center}
  \hspace*{-5mm}
    \includegraphics[width=0.85\hsize]{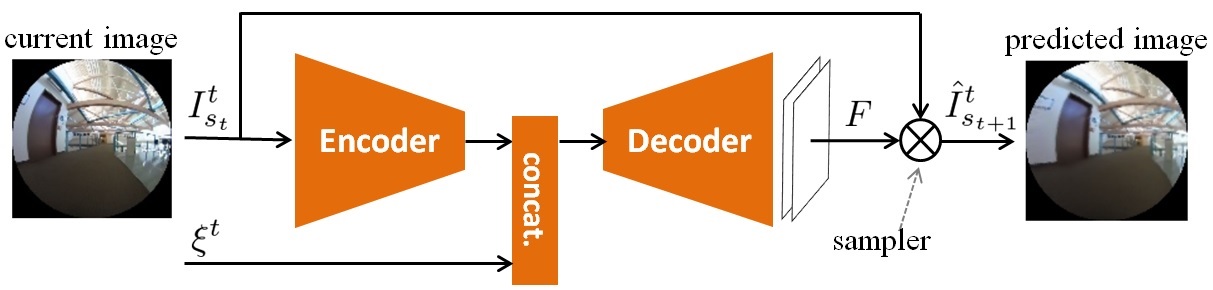}
  \end{center}
  	\caption{\footnotesize Static Transformation Network (SNet) structure. The input is an RGB image, $I_{s_t}^{t}$ from the robot's fish-eye camera at time $t$ and a camera pose $s_t$ , and a virtual future twist velocity, $\xi^t$. The output is a predicted image from the future robot pose, $I_{s_{t+1}}^{t}$. The network decides how to change (sample from) the current image to generate the virtual image based on a flow field, $F$.}
	\label{f:stnet}
	\vspace{-4mm}
\end{figure}

Figure~\ref{f:stnet} shows the network structure of SNet. 
SNet uses an image from a camera pose $s_t$ and a virtual velocity $\xi^{t}$ to predict an image from a different camera pose $s_{t+1}$ (changes in the spatial domain).
The architecture is based on the encoder-decoder architecture (ED).
Our encoder-decoder has two special characteristics: 1) the virtual velocity input is concatenated to the low dimensional image representation to realize the spatial transformation before the decoding phase, and 2) the output of the decoder is a 2D flow field image ($F$) that is used to sample the original input images and generate the predicted future image. SNet generates sharper images than the classical ED architectures because the sampling procedure reuses original pixels of the input image using the internal flow field representation (see Fig.~\ref{f:pimage}).

\begin{figure}[t]
  \begin{center}
  \hspace*{-5mm}
    \includegraphics[width=0.85\hsize]{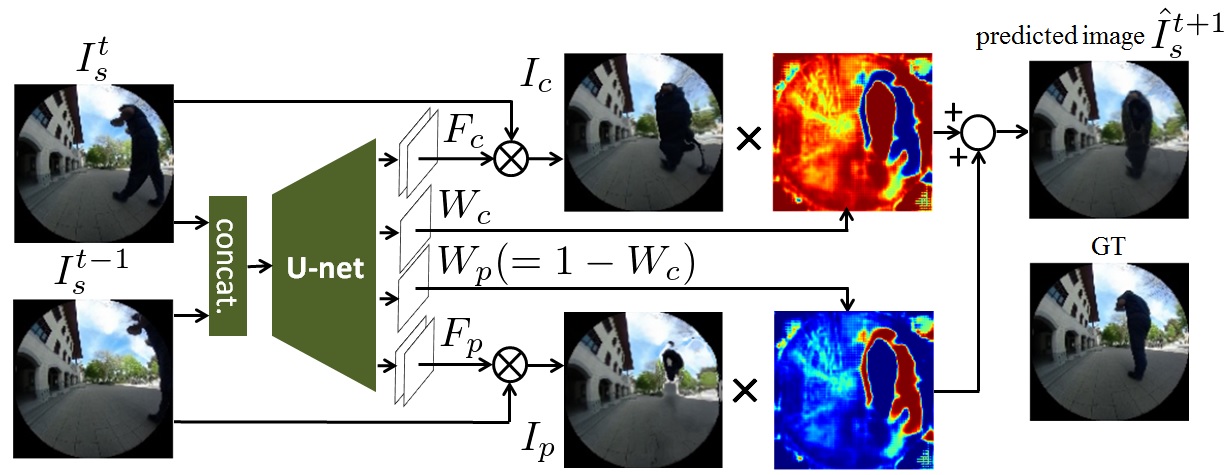}
  \end{center}
  	\caption{\footnotesize Dynamic Transformation Network (DNet) structure. The input are two RGB images at time $t$ and $t-1$ at location $s$. The output is a predicted image in the next time step, $t+1$, from the same location considering dynamic moving objects. Our network decides how to sample from the images at previous and current time steps based on the generated flow fields $F_{p}$ and $F_{c}$, and how to alpha-blend the samples based on the probabilistic selections masks $W_{p}$ and $W_{c}$. Both sampled images and the probabilistic masks are depicted (red indicates high weight for the merge, blue color indicates low weight)}
	\label{f:dnet}
	\vspace{-4mm}
\end{figure}
%

\paragraph*{\textbf{Dynamic Transformation Network (DNet)}}

Figure \ref{f:dnet} depicts the architecture of the DNet.
DNet takes as input two images (real or virtual) acquired from the same camera pose ($s$) in consecutive time steps and synthesizes a virtual image in the next time step. The synthesized image accounts for the changes due to the motion of the dynamic objects in the scene (changes in the temporal domain).

To synthesize the image, DNet generates four intermediate representations: two 2D flow field images, $F_{c}$ and $F_{p}$, to sample pixels from current and previous images respectively, and two 1D probabilistic selection masks, $W_{c}$ and $W_{p}$, to weight the contribution of the samples from the current and previous images in a final alpha-blend merge. We use a softmax function to generate $W_p$ and $W_c$ that satisfies $W_p(u,v) + W_c(u,v) = 1$ for any same image coordinates $(u,v)$. The intermediate representation is generated by a U-net architecture~\cite{ronneberger2015u} that has been successfully applied before to other virtual image synthesis tasks by~\citet{pix2pix2017}.

We use two consecutive images in DNet for two reasons: 1) a single image does not have the pixel information of the parts of the environment occluded by the dynamic object, and 2) a single image does not contain motion information of the dynamic object.
Using two images we can acquire pixel information behind the dynamic obstacles and also understand their motion behavior as illustrated in Fig.~\ref{f:dnet}.

\paragraph*{\textbf{Dynamic-Scene View Synthesis Architecture (VUNet)}}

Figure \ref{f:overview_net} shows the overall structure of proposed approach, VUNet for view synthesis in dynamic environments by combining SNet and DNet. 
VUNet is composed of two steps. In the first step, the method applies SNet on the previous and current images to predict the virtual images of current and previous time step as they would be seen from robot's next pose, $s_{t+1}$. This step is shown as an orange arrows in Fig.~\ref{f:overview_net:a} and as a light orange block in Fig.~\ref{f:overview_net:b}.
Note that, after this step the difference between the two predicted virtual images is expected to be caused only by the motion of the dynamic objects. 

In the second step, our method feeds the two virtual images predicted by the previous step into DNet. DNet predicts the pixel changes caused by the motion of the dynamic objects and generates the final synthesized image: an image at the new pose $s_{t+1}$ at the future time step $t+1$.
This step is shown as a green arrow in Fig.~\ref{f:overview_net:a} and a light green block in Fig.~\ref{f:overview_net:b})

By combining SNet and DNet, VUNet can predict future images that satisfy both static and dynamic changes in an environment caused by robot and dynamic objects movements.

\subsection{Future Traversability Estimation}
\label{ssec:fte}
\begin{figure}[t]
  \begin{center}
    \includegraphics[width=0.85\hsize]{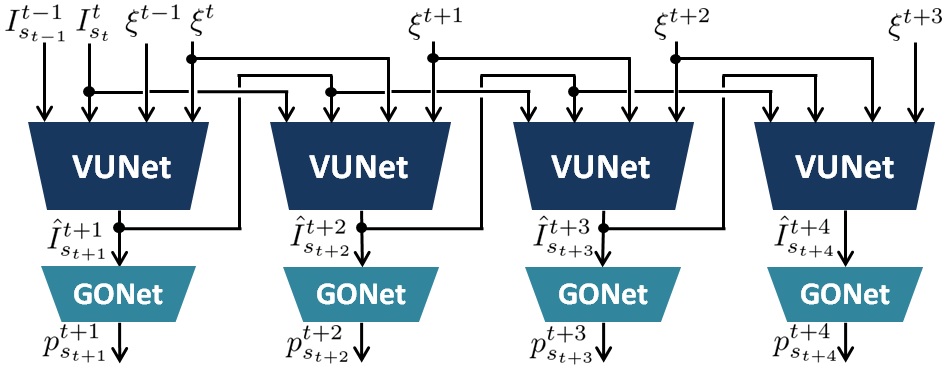}
  \end{center}
  	\caption{\footnotesize{System for multi-step future image prediction and traversability estimation. The input to our system are the current and previous images, the last velocity twist command $\xi^{t-1}$ and a series of virtual velocity twists $\xi^{t+i}$. Each block of VUNet represent the system of Fig.~\ref{f:overview_net:b}. They generate predicted virtual images at next time steps. These images are 1) passed to our previously presented method GONet~\cite{hirose2018gonet} to estimate the traversable probability $p^{t+i}_{s_{t+i}}$, and 2) passed as input to the next VUNet block to predict the next image.}}
	\label{f:fig_msg}
	\vspace{-4mm}
\end{figure}
We now show how we apply VUNet to the estimation of traversability in future steps. This process is composed of two steps: First, we use the current and previous acquired images as well as the last robot's velocity ($\xi^{t-1}$) and a virtual velocity ($\xi^{t}$) to generate a first predicted future image (most left VUNet block in Fig.~\ref{f:fig_msg}).
The generated virtual image is passed to our previously proposed GONet architecture~\cite{hirose2018gonet} -- an RGB-based method that estimates the traversable probability of a scene image. The virtual image is also passed to the next VUNet block to generate a virtual image at the next time step.
By repeating this process for multiple time steps, our approach is able to estimate the traversable probability at $n$ consecutive future time steps, assumed $n$ future virtual velocities. The general approach to estimate traversability in future steps is depicted in Fig.~\ref{f:fig_msg}. 

\section{Experimental Setup}
\label{sec:setup}

\subsection{Implementation}
\label{sec:implementation}
\textbf{SNet:} In SNet we use as network structure a regular encoder-decoder (ED) similar to the network used by~\citet{pix2pix2017} but without the skip connections and the dropout.
The input to the encoder is a three channel (RGB) $128\times128$ image. The output of the encoder is a 512 dimensional feature vector that we concatenate with the two dimensional velocity vector $\xi^t$ to feed it to the decoder. The decoder generates a flow field image $F$ of size $2\times128\times128$ that we use for bilinear sampling.
The synthesized image resulting from SNet is a three channel (RGB) $128\times128$ image. 

\textbf{DNet:} The DNet is based on the network architecture of U-net~\cite{ronneberger2015u} as used in~\cite{pix2pix2017}. Different to~\cite{pix2pix2017}, in DNet the inputs are two 3 channel (RGB) $128\times128$ images and the outputs (intermediate representation) are two  $2\times128\times128$ flow field images and two $128\times128$ probabilistic masks. The final output of DNet is a three channel (RGB) $128\times128$ image resulting from an alpha-blend process.

The GONet network used for future traversability estimation is a pretrained network as explained in~\cite{hirose2018gonet}. All networks are implemented in Chainer~\cite{tokui2015chainer} and our sampling period (time between consecutive steps) is \SI{0.33}{\second}.

\subsection{Training}
\label{sec:dataset}

To train the different components of our model we will need two different types of data: data where the robot moves in a static environment to train SNet, and data that includes dynamic objects without the robot motion to train DNet.

To train SNet we use the \emph{GO Stanford 2} (GS2) dataset presented in~\cite{hirose2018gonet}. GS2 contains 16 hours and 42 minutes of videos from 27 campus buildings acquired from a Ricoh THETA S fisheye camera on a teleoperated Turtlebot2 robot, and the velocity commands to this robot. Even though some few sequences in GS2 include dynamic objects, their number is very small and they do not affect the training process of SNet. We randomly flip the image and invert the angular velocity for the data augmentation to avoid overfitting in the training process.

To train DNet we record new data from a constant robot position observing dynamically moving objects (humans, vehicles, \ldots). We maintain the robot at a fixed position to have only image changes caused by the motion of the dynamic objects. 
We use the same robot and camera to record 4 hours and 34 minutes (47730 images) of videos at 46 different points in 23 different indoor and outdoor environments. We also make this new dataset ``GO Stanford 3'' (GS3) available to the community~\footnote{\url{http://svl.stanford.edu/projects/vunet/}}. 

We could train directly DNet on pairs of current and previous images from GS3. However, as explained in Sec.~\ref{ssec:dsvs}, the input to DNet in our method is the output of SNet. SNet often generates some small disturbances. To train DNet on data with these disturbances we preprocess GS3 images passing them through a trained SNet with a small random velocity perturbation $\epsilon_{\xi}$, uniformly distributed between $\pm0.05$ in all dimensions.

To train both SNet and DNet we use data from separate locations (i.e. different buildings or campus areas) for training, test, and validation. This way, the evaluation on test and validation assesses how well our method generalizes to completely new environments. This location-based splits the data to $70\%$ training, $15\%$ test, and $15\%$ validation.

We iteratively train all networks with a batch size of 80 using Adam optimizer~\cite{kingma2014adam} and with a learning rate of 0.0001. Our networks are trained by minimizing the L1 norm. For real-world experiments, our proposed system for view synthesis is implemented on a robot with a laptop equipped with a Nvidia Geforce GTX 1070 GPU that allows us to maintain 3 fps of constant computation time.

Additionally, we collected videos of teleoperated robot trajectories in dynamic environments with humans to use for the evaluation of our view synthesis and future traversability estimation methods. We recorded 26 minutes of videos in six different environments and include it as part of GS3 and more than one hour of highly dynamic environments.

\section{Experiments}
\label{sec:experiments}

We conducted two sets of experiments. In the first set we evaluate quantitatively the performance of our view synthesis method and our system for future traversability estimation for mobile robots in dynamic environments. In the second set of experiments we evaluate all methods qualitatively.

\subsection{Quantitative Analysis}

\textbf{DNet:} First we evaluate the performance of only DNet on the test data of GS3 where the robot is static. In this scenario it is not necessary to use SNet, because the camera viewpoint does not change. Hence, we can evaluate DNet separately. We compare DNet to several baselines: DNet variants using a regular encoder-decoder (ED) without path skips (instead of the U-net architecture), without multi-image merge, and a variant using an extrapolation based on optical flow (OF) between previous and current image using FlowNet~\cite{dosovitskiy2015flownet}. We report mean L1 norm (pixel difference) and structural similarity (SSIM~\cite{wang2004image}) between the generated images and the ground truth future images of the test data.

Table~\ref{tab:DNet} depicts the result of our quantitative analysis on DNet and the comparing baselines. In this table ``+S'' indicates sampling, and ``+M'' indicates multi-image merge (last step of DNet). 
The U-net architecture used in DNet outperforms the methods using a regular encoder-decoder(ED). 
Also, our multi-image merge (M) approach leads to better results (lower L1 and higher SSIM) than the single image approaches. 
Moreover, our method also outperforms the optical flow (OF) based baseline.

\begin{table}[t!]
  \centering
  \caption{Evaluation of DNet}
  \label{tab:DNet}
  \begin{tabular}{|l|c|c|c|c|c|}\wcline{1-6}
       & OF~\cite{dosovitskiy2015flownet} & ED+S & U-net+S & ED+S+M & U-net+S+M \\ \wcline{1-6} 
       L1 & 0.146 & 0.135 & 0.119 & 0.113 & \textbf{0.104} \\ \cline{1-6}
       SSIM & 0.649 & 0.698 & 0.710 & 0.703 & \textbf{0.727} \\ \cline{1-6}
	\end{tabular}
	\vspace{-4mm}
\end{table}

\textbf{Dynamic-Scene View Synthesis:} We evaluate the performance of our complete view synthesis method, VUNet for dynamic environments in test data with i) static environments in GS2, ii) dynamic environment with fixed robot position from GS3, and iii) dynamic environment with moving robot from GS3. This last group of sequences is especially challenging because this type of data has not been seen during SNet and DNet training. We compare multiple structures for SNet and DNet using regular encoder-decoder versus U-net architecture, training with and without GAN, and optical flow. We report mean L1 norm (pixel difference) and structural similarity (SSIM) between the generated images and the ground truth future images in the test data. 

Table \ref{tab:analysis} shows the quantitative results of the baseline methods and the ablation study on the proposed method to generate virtual images (SNet+DNet). For each model and type of data the first value indicates the mean pixel L1 difference (smaller is better) and the second value indicates the structural similarity, SSIM (larger is better).
The six rows from (b) to (g) depict the results of synthesizing images using \textbf{only} different variations of the SNet architecture. Without sampling, U-net (row (d)) outperforms the regular encoder-decoder (ED, row (b)). However, with sampling the encoder-decoder architecture (row (f)) improves the performance, especially in our application scenario, dynamic environments with robot motion (third column). 
We also observed that using a GAN training does not improve the results (rows (c) and (e)).

\begin{table*}[h]
  \centering
  \caption{Evaluation of View Synthesis and Traversability Estimation}
  \centering
  \resizebox{1.8\columnwidth}{!}{%
  \label{tab:analysis}
  \begin{tabular}{|llll|c|c|c|c|c|}\wcline{1-9}
       \hspace{0mm} Models: & & & & GS2 & GS3 & GS3 & \multicolumn{2}{c|}{Trav. Accuracy} \\
       \hspace{2mm} SNet Variants & + & DNet Variants & & Static Environment & Dynamic env. wo robot motion & Dynamic env. w/ robot motion & GS2 \& 3 & Ped. DS \\ \wcline{1-9} 
       (a) Kinect & & & & - & - & - & 0.818 & 0.735 \\ \cline{1-9}
       (b) ED\cite{tatarchenko2016multi} &+& - & & 0.117 \, / \, 0.556 &  0.225 \, / \, 0.395 & 0.151 \, / \, 0.501 & 0.690 & 0.620 \\ \cline{1-9}
       (c) ED+GAN\cite{larsen2015autoencoding} &+& - & & 0.147 \, / \, 0.468 & 0.253 \, / \, 0.333 & 0.188 \, / \, 0.400 & 0.660 & 0.540 \\ \cline{1-9}
       (d) U-net\cite{ronneberger2015u} &+& - & & \textbf{0.064} \, / \, \textbf{0.779} & 0.148 \, / \, \textbf{0.698} & 0.115 \, / \, 0.644 & 0.920 & 0.735 \\ \cline{1-9}
       (e) U-net+GAN &+& - & & 0.069 \, / \, 0.752 & 0.148 \, / \, \textbf{0.698} & 0.124 \, / \, 0.602 & 0.920 & 0.675 \\ \cline{1-9}
       (f) ED+S\cite{zhou2016view} &+& - & & \textbf{0.065} \, / \, \textbf{0.777} & 0.155 \, / \, 0.672 & 0.116 \, / \, 0.647 & 0.947 & 0.777 \\ \cline{1-9}
       (g) U-net+S &+& - & & 0.067 \, / \, 0.765 & 0.158 \, / \, 0.663 & 0.117 \, / \, 0.642 & 0.945 & 0.770 \\  \cline{1-9}
       (h) U-net+S &+& OF & & 0.086 \, / \, 0.706 & 0.155 \, / \, 0.607 & 0.143\, / \, 0.548 & 0.905 & 0.822 \\ \cline{1-9}
       (i) U-net+S &+& ED+S+M & & 0.068 \, / \, 0.761 & 0.123 \, / \, 0.668 & 0.108 \, / \, 0.647 & 0.945 & 0.797 \\ \cline{1-9}
       (j) U-net+S &+& U-net+S+M & & 0.068 \, / \, 0.765 & 0.116 \, / \, 0.686 & 0.105 \, / \, 0.653 & 0.937 & 0.810 \\ \cline{1-9}
       (k) ED+S &+& OF & & 0.092 \, / \, 0.680 & 0.158 \, / \, 0.594 & 0.594\, / \, 0.529 & 0.905 & 0.817 \\ \cline{1-9}       
       (l) ED+S &+& ED+S+M & & 0.068 \, / \, 0.766 & 0.123 \, / \, 0.686 & 0.110 \, / \, 0.644 & 0.937 & 0.800 \\ \cline{1-9}       
       (m) \textbf{ED+S} &+& \textbf{U-net+S+M} & \textbf{(VUNet)} & \textbf{0.065} \, / \, \textbf{0.776} & \textbf{0.113} \, / \, \textbf{0.698} & \textbf{0.104} \, / \, \textbf{0.657} & \textbf{0.950} & \textbf{0.830} \\ \cline{1-9}
    \end{tabular}
    }
\end{table*}

The last six rows (from (h) to (m)) depict the result of synthesizing images combining different variants of SNet and DNet, including our proposed method (row (m)). Our proposed method, VUNet is one of the best three methods in all scenarios. Specifically, in our application domain, dynamic scenes with robot motion, our method outperforms all baselines. 
We observe that the usage of DNet does not improve the results on static environments (first column). In these scenarios it is simpler to use only SNet. However, as expected, SNet alone fails in scenarios with dynamic objects.

\textbf{Future Traversability Estimation:} We evaluate the accuracy of the estimation of traversability based on the images generated by our view synthesis and the previously proposed baselines. We randomly sample images from GS2 and GS3 and hand-label them until we collect 200 traversable and 200 untraversable images. The untraversable images are images just before the robot collides or falls. We take the two previous images to each selected image and use them together with the ground truth commanded velocity to predict the selected image. We feed then the generated image to GONet~\cite{hirose2018gonet} to calculate the future traversable probability. If the probability is over a threshold $p_{min} = 0.5$ we label the image as traversable, otherwise we label it as untraversable. We estimate the accuracy of the traversability estimation by comparing the predicted and the manually assigned (ground truth) traversability labels.

The left side of the last column of Table~\ref{tab:analysis} depicts the results of this quantitative evaluation on the future traversability estimation in GS2 and GS3. Feeding the images generated by VUNet for view synthesis (row (m)) to GONet yields the highest accuracy for the traversability estimation. The higher accuracy is the result of a higher quality in the predicted images.
We note that the accuracy in the traversability estimation of some variants without DNet is also high (e.g. ED+S, row (f), and U-net+S, row (g)). This is an artifact caused by the distribution of our evaluation data: even though we sample 100 evaluation images from GS3 depicting dynamic objects, they are usually far and failing to predict their changes in the image do not usually affect the traversability estimation. 

In order to evaluate more clearly the benefits of using our DNet component, we collected an additional dataset of about one hour (\textit{Ped. DS} in Table~\ref{tab:analysis})
where the dynamic obstacles (pedestrians) and the robot often cross their trajectories.
We compare our future traversability estimation method to the baselines and list their accuracy in the right side of the last column in Table~\ref{tab:analysis}. Our method achieves the highest accuracy and shows a clear quantitative advantage against baseline methods without DNet, i.e. not accounting for the motion of the dynamic obstacles.

We also compared VUNet to a baseline using depth images from a Kinect sensor included also in GS2, GS3 and the pedestrian dataset.
We turn the Kinect sensor pointing forward into a proximity sensor and develop a baseline that indicates that an area is untraversable if there are obstacles closer than a distance threshold. We determined the optimal threshold as the threshold that leads to the maximum accuracy in the validation set.
The accuracy of the Kinect-based baseline is listed in the first row of Table~\ref{tab:analysis}. The baseline using depth images performs worse than our RGB-based method because the Kinect images contain noise due to reflections in mirrors, glass walls, and missing points in dark objects. Also, the traversability estimation using Kinect do not consider the motion of the pedestrian in the future step, which leads to the poor performance in the pedestrian dataset.

Additionally, we evaluated our proposed approach, VUNet for future traversability to predict two, three and four steps obtaining $91.3\%$, $89.3\%$ and $88.0\%$, respectively. The decreased accuracy is caused by the more difficult predictions in longer horizons. Considering safety applications we evaluated the accuracy using a more conservative traversability threshold of $p_{min} = 0.7$ decreasing the amount of non-predicted untraversable future steps (the riskiest case) to less than $6\%$ in one to four future steps.

\begin{figure*}[t]
  \begin{center}
  \hspace*{-5mm}
    \includegraphics[width=0.82\hsize]{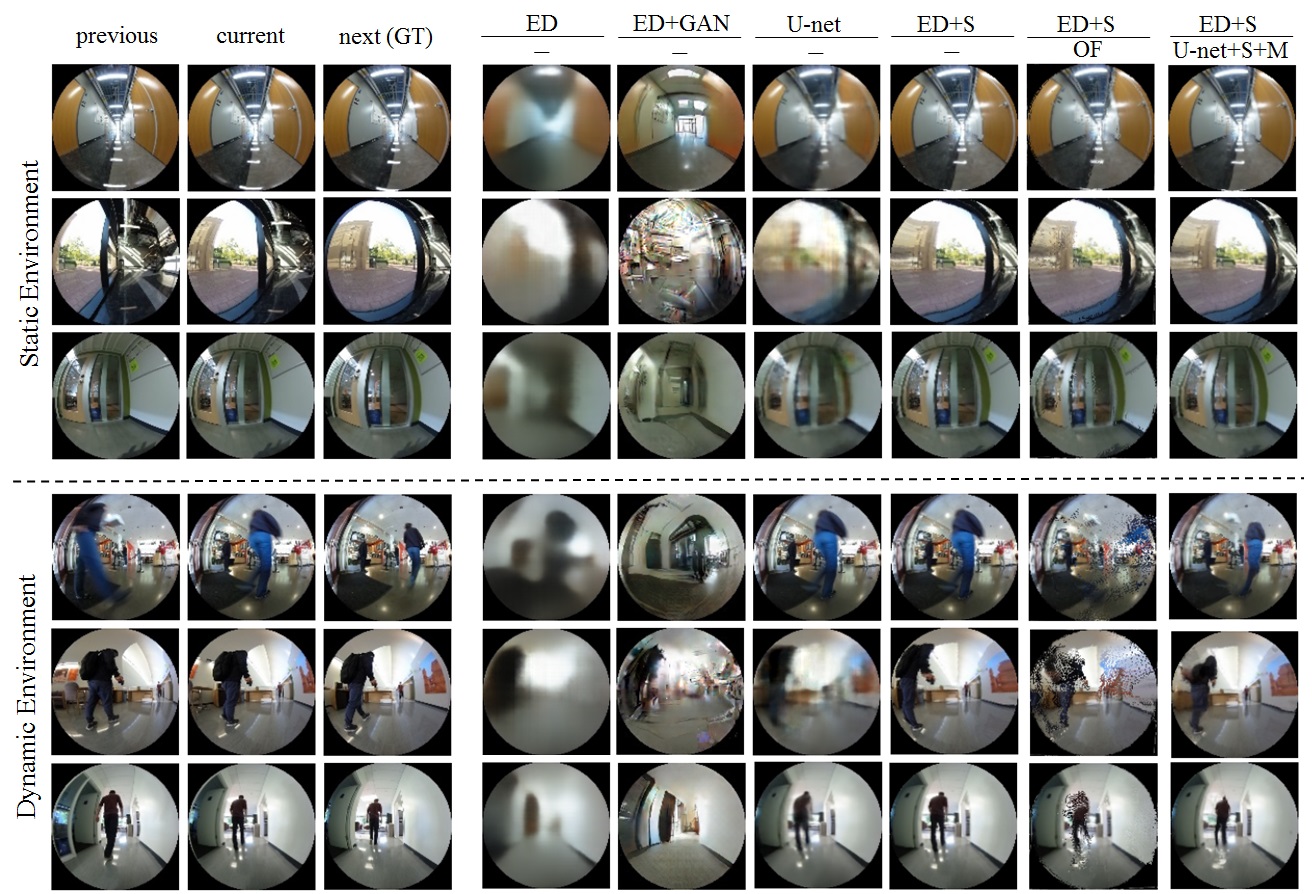}
  \end{center}
  	\caption{\footnotesize{Predicted images in static (first three rows) and dynamic (last three rows) environments. From left to right: previous, current and future (ground truth for view synthesis, GT) images, predicted images from baselines (SNet and DNet variants), and predicted images from VUNet (most right)}}
	\label{f:pimage}
\end{figure*}
\subsection{Qualitative Analysis}

In the second set of experiments we evaluate qualitatively the results of our dynamic-scene view synthesis, VUNet and traversability estimation approaches. 

\textbf{Dynamic-Scene View Synthesis:} First, we compare the generated images from our method and the baselines methods side by side (see Fig.~\ref{f:pimage}). 
The first three columns show the previous, current, and future (ground truth for the synthesis, GT) RGB images as viewed by the robot. 
We observe that the GAN training improves the sharpness of the blurred predicted image from the encoder-decoder (ED and ED+GAN, 4th and 5th columns).
However, while being sharper, some of the generated images by ED+GAN do not resemble much the real image (e.g. 2nd and 5th row of 5th column).
U-net (6th column) can generate very clear images when the current image is similar to the future image, but it does not perform as well when it has to predict dynamic obstacles.
Similarly, all baseline methods without DNet are not able to predict the appearance changes due to moving objects (e.g. humans). We observe that the location of humans in the predicted images without DNet is same as the future predicted image (three last rows).
We can also see that, ED+S+OF can not predict accurately the human movement: there are speckle patterns in the predicted images. This is because the errors in SNet cause wrong extrapolations with OF.

In contrast, our method, VUNet (SNet+DNet) is able to predict the image changes due to both robot pose changes and motion of the dynamic objects (i.e. humans).
For example, in the scene shown in the 5th row, the robot is turning to the right side while a human is crossing by. Surprisingly, even the unseen part of the picture on the right side wall in the future image can be correctly constructed in the predicted image.
Additionally, the human is correctly moved towards the right in the predicted image (a failure in ED+S). 

In the last scene (last row), both the robot and the human are moving forward in the corridor. While several methods can correctly predict that the static parts of the environment (e.g. the door) will appear closer to the robot, only our method predicts that the human, which is faster than the robot (as can be observed from previous and current images), should be farther away in the predicted image.

\textbf{Future Traversability Estimation:} To evaluate qualitatively our method for future traversability estimation we propose two applications based on it for assisted teleoperation: early obstacle detection or multi-path future traversability estimation. These methods are implementations of the system depicted in Fig.~\ref{f:fig_msg} with different ways of generating virtual future robot velocity commands. 

\textit{Early obstacle detection:} In this application, the teleoperator uses a joypad to control the robot and gets audio input (warning) or emergency stops from the proposed system.
To predict the images and the traversability in the future our method assumes that the upcoming robot commands will be the robot's maximum linear velocity and a constant angular velocity $\xi=(v_{max}, v_{max}/r_c)$. With this assumption our method assumes the riskiest possible future.
The robot's maximum linear velocity is \SI{0.5}{\meter\per\second}, and $r_c$ is the turning radius last used by the teleoperator calculated as $r_c=v_c/\omega_c$, where $v_c$ and $\omega_c$ are the teleoperator's last commands.
A safety alarm is fired when the traversable probability for the third ($p_{s_{t+3}}^{t+3}$) or fourth ($p_{s_{t+4}}^{t+4}$) time steps in the future are less than 0.5.
Additionally, an emergency stop interrupts the current teleoperation of the robot if the traversable probability of the current state ($p_{s_{t}}^{t}$), next ($p_{s_{t+1}}^{t+1}$), or second next steps ($p_{s_{t+2}}^{t+2}$) are less than 0.5.

\begin{figure}[t]
  \begin{center}
    \includegraphics[width=0.85\hsize]{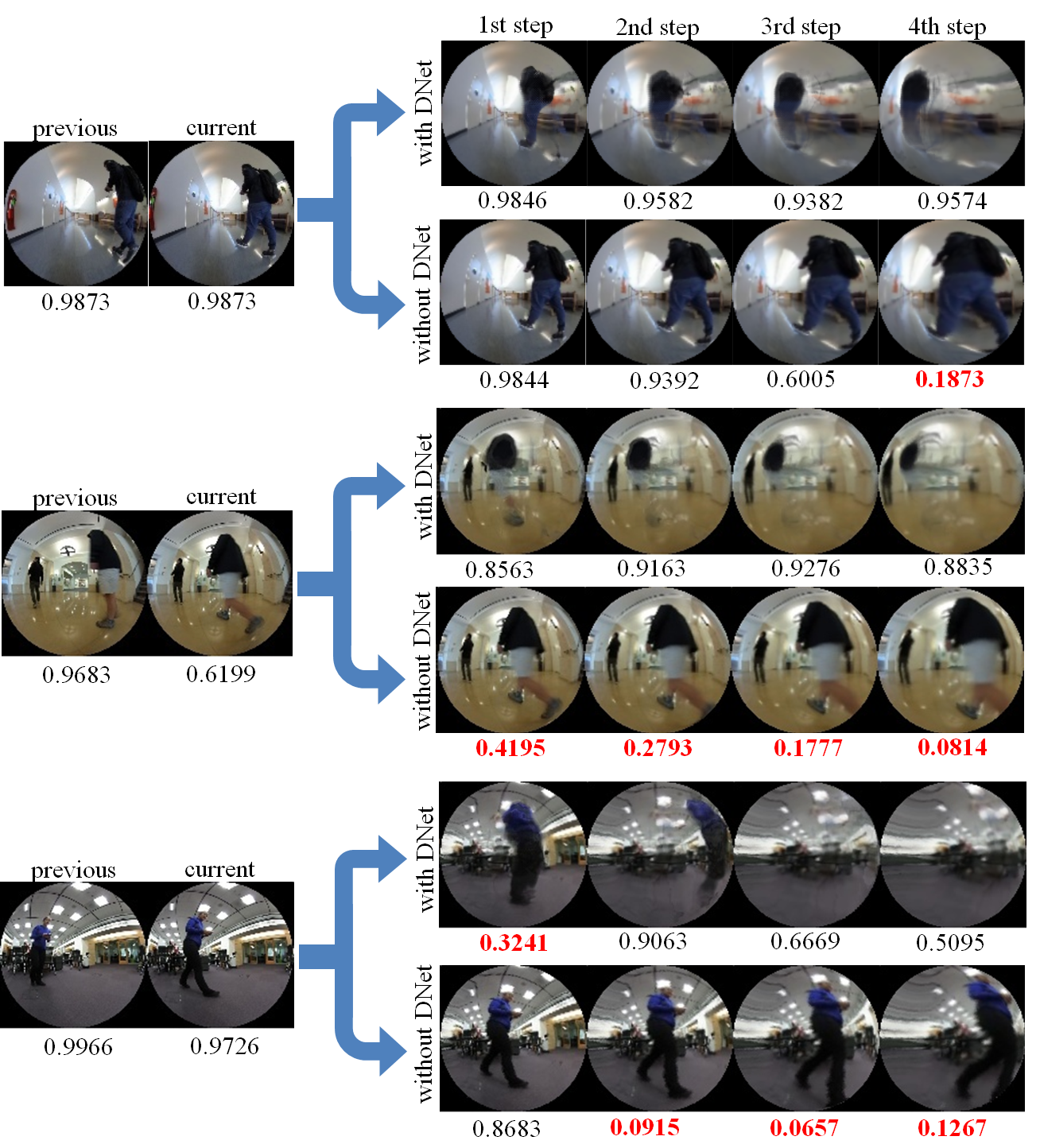}
  \end{center}
  	\vspace{-4mm}
	\caption{\footnotesize{Application of our future traversability estimation for early obstacle detection in three dynamic environment examples. The inputs to our application system are the previous and current images, and the last teleoperation command. The system predicts the images at four consecutive time steps and estimates the traversable probability, depicted under each image. Red probabilities indicate values under 0.5. We show the different predictions using DNet(VUNet) or without DNet (only SNet.)}}
	\label{f:mstep}
	\vspace{-6mm}
\end{figure}

Fig.~\ref{f:mstep} shows three examples of our application for early obstacle detection of cases in dynamic environment with moving obstacles (pedestrians).
The traversable probability of each image is shown under each image by applying GONet to each image generated with VUNet.
We compare the results of future traversability estimation based on images using only SNet or our proposed method, VUNet (SNet+DNet).
In the predicted images without DNet, the changes due to the motion of the human cannot be predicted. Therefore, the model without DNet wrongly estimates the traversability assuming a non-moving pedestrian. Our proposed approach using DNet predicts the motion of the human in the image, which leads to a more accurate prediction of the future traversability.
Additional qualitative results can be seen in our supplementary video.
Our proposed application for early obstacle detection correctly estimates the traversable probability in the future and indicates this to the teleoperator with warning signals and emergency stop commands.

\textit{Multi-path traversability estimation:} For the future traversability estimation for multiple paths, we propose to apply our system of Section~\ref{ssec:fte} in the way depicted in Fig.\ref{f:fig_msg}, top. The system generates virtual velocities for five different paths around the robot, predicts the images using our scene view method and calculate the traversability for each of the paths.
To generate the virtual robot velocities $\xi^{t+i}, i \in {1 \cdots 4}$ our system assumes a constant maximum linear velocity, $v_{max} = \SI{0.5}{\meter\per\second}$ and five different angular velocities multiple of $\omega_{0}=\SI{0.5}{\radian\per\second}$: $\omega_{LL}=2\omega_{0}$, $\omega_{L}=\omega_{0}$, $\omega_{C}=0$, $\omega_{R}=-\omega_{0}$, and $\omega_{RR}=-2\omega_{0}$.
We ask a teleoperator to navigate the robot in different scenarios and collect the multiple path traversability predictions. 

\begin{figure}[t]
\begin{center}
\begin{subfigure}[b]{0.49\textwidth}
\hspace{1.5cm}
\includegraphics[width=0.75\hsize]{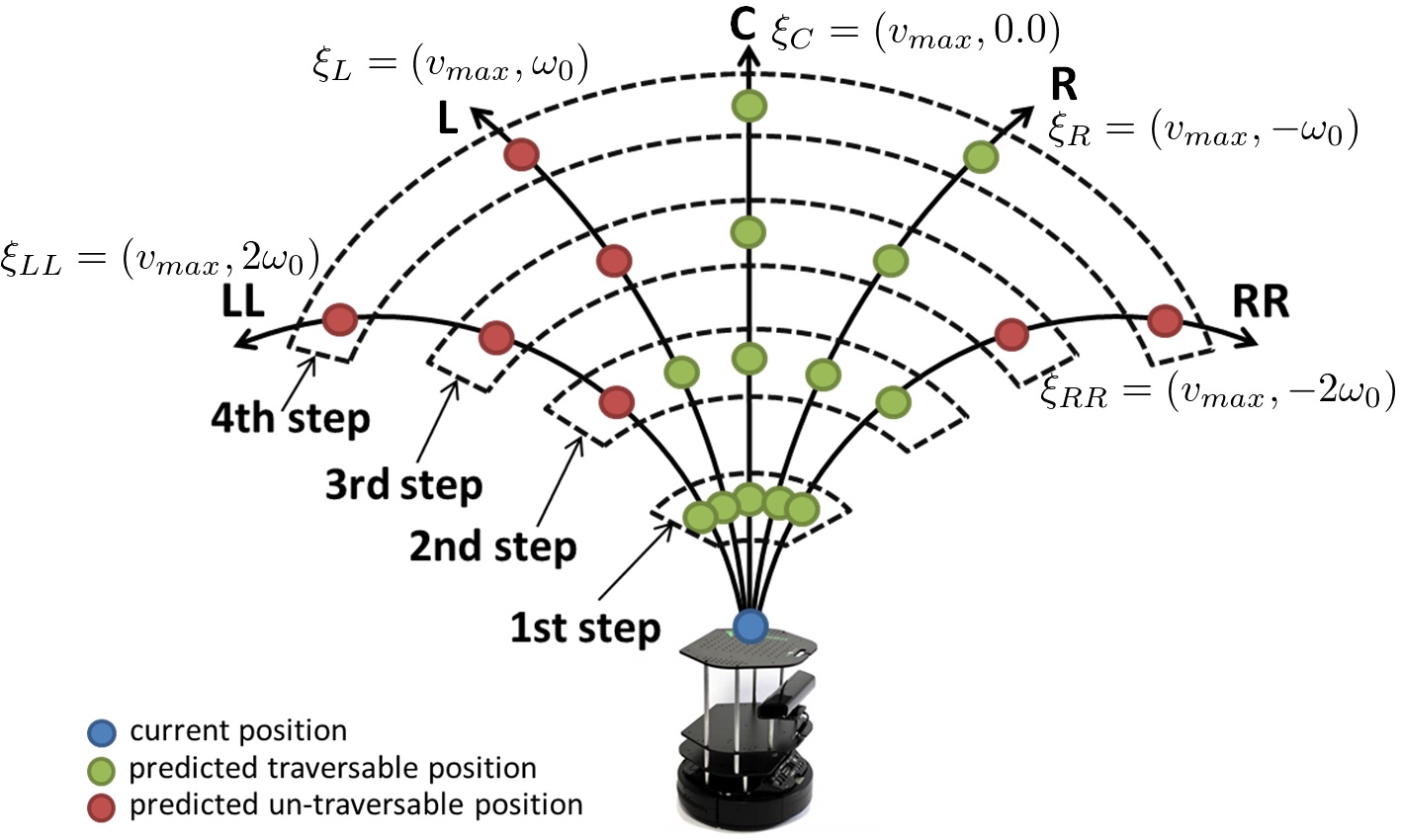}
\label{f:pmimage:a}
\end{subfigure}
\begin{subfigure}[b]{0.49\textwidth}
\centering
\includegraphics[width=0.8\hsize]{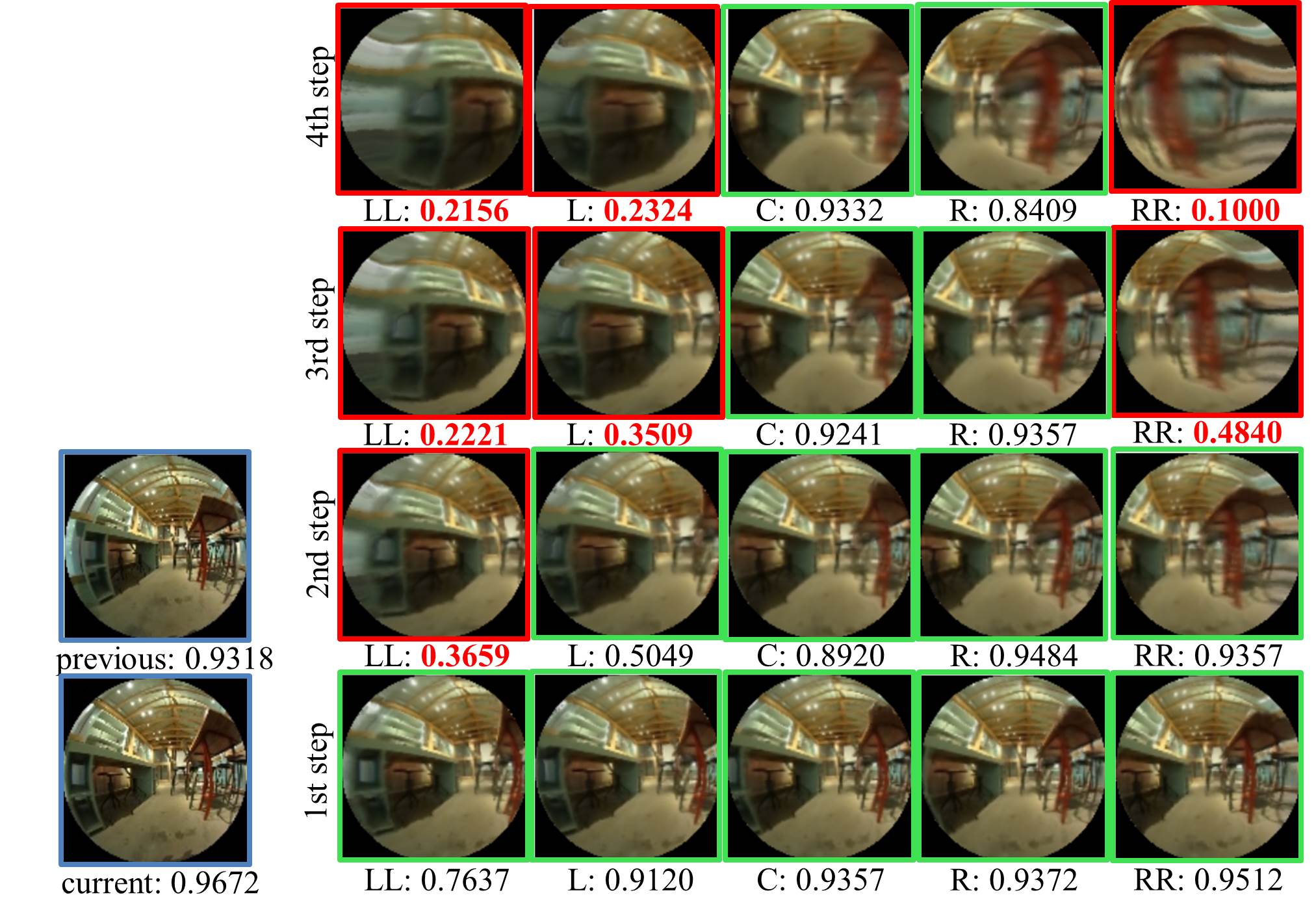}
\label{f:pmimage:b}
\end{subfigure}
\end{center}
\caption{\footnotesize{Application of our multi-path traversability estimation system for assisted teleoperation. 
Top: Spatial diagram of the five paths where we predict traversability around the robot. We estimate traversability in a most left (LL), left (L), central (C), right (R), and most right (RR) paths. Bottom: the input previous and current images (left) and the generated images (right) with associated traversable probability. The robot image is cited from \cite{turtlebot}.}}
\label{f:pmimage}
\end{figure}

Figure \ref{f:pmimage} shows an example of our multi-path traversability estimation. 
The figure shows previous and current images (left side) as well as the predicted images on each path on the test set (bottom). The traversable probabilities are shown under each image.
In this example, the robot is moving in a narrow corridor with tables and chairs on both sides.
Our method can correctly predict the safe path in front of the robot based on the synthesized future images. Additional qualitative examples of this application are included in our supplementary video.

\section{Limitations and Future Work}
\label{sec:limitations}

The quantitative and qualitative analysis pointed out some limitations but we don't deem them severe for our applications. The quality of the synthesized images decreases for longer time horizon predictions. This affects the accuracy of the future traversability estimation. The degradation of future image predictions is caused by two main factors: 1) the part of the environment that is now visible was not visible in the images we used to synthesize the view (e.g. due to large occlusions or abrupt rotations), and 2) the dynamic objects present a complex motion pattern (e.g. different parts of the human body like legs and arms).
However, even in these scenarios, the quality of the generated images is good enough to predict with high accuracy the future traversable probability for the spaces around the mobile robot.
To alleviate further these effects we will explore in future work methods to reflect the uncertainty on the predictions, both due to odometry errors and due to non-deterministic dynamic obstacle motion\cite{sadeghian2018sophie} We also plan to combine our approach with a vision-based target navigation into a full autonomous navigation system that avoids obstacles and reaches a target destination.

\section{Conclusion}
\label{sec:conclusion}

In this paper we propose a novel dynamic-scene view synthesis method for robot visual navigation using an RGB camera, VUNet. We show the proposed method is applicable for future traversability estimation. Our view synthesis method predicts accurate future images given virtual robot velocity commands. Our method is able to predict the changes caused both from the moving camera viewpoint and the dynamically moving objects. Our synthesized images outperform both quantitatively and qualitatively the images generated by state of the art baseline methods. We use the synthesized images to predict traversability in future steps for multiple paths and show its application to assisted teleoperation scenarios.

\balance
\bibliographystyle{IEEEtranN}
\vskip-\parskip
\begingroup
\footnotesize
\bibliography{reference}
\endgroup
\end{document}